\documentclass[journal]{IEEEtran}
\usepackage{epsfig}
\usepackage{cite}
\usepackage{graphicx}
\usepackage{subfigure}
\usepackage{amsmath}
\usepackage{amssymb}
\usepackage{verbatim}
\usepackage{algorithm}
\usepackage{algorithmic}
\usepackage{booktabs}
\usepackage{multirow}
\usepackage{mathtools}
\usepackage{breqn}
\usepackage{dirtytalk}

\ifCLASSINFOpdf
\else
\fi
\begin{document}
%
\title{Interactive Medical Image Segmentation using Deep Learning with Image-specific Fine-tuning}
%
%
%

\author{Guotai Wang, 
		Wenqi Li,
        Maria A. Zuluaga,
        Rosalind Pratt, 
        Premal A. Patel,
        Michael Aertsen,
        Tom Doel,
        Anna L. David, 
        Jan Deprest,
        S\'ebastien Ourselin, 
        and Tom Vercauteren
\thanks{G. Wang, W. Li, M A. Zuluaga, R. Pratt, P. Patel, T. Doel, S. Ourselin and T. Vercauteren are with the Department
of Medical Physics and Biomedical Engineering, and Wellcome EPSRC Centre for Interventional and Surgical Sciences (WEISS), University College London, London,
WC1E 6BT, UK. T. Vercauteren is also with KU Leuven. e-mail: guotai.wang.14@ucl.ac.uk.}
\thanks{A L. David is with WEISS and Institute for Women's Health, University College London. M. Aertsen and J. Deprest are with University Hospitals KU Leuven. J. Deprest is also with WEISS.}
\thanks{This work has been submitted to the IEEE for possible publication. Copyright may be transferred without notice, after which this version may no longer be accessible.}}
\maketitle

\begin{abstract}
Convolutional neural networks (CNNs) have achieved state-of-the-art performance for automatic medical image segmentation. However, they have not demonstrated sufficiently accurate and robust results for clinical use. In addition, they are limited by the lack of image-specific adaptation and the lack of generalizability to previously unseen object classes. To address these problems, we propose a novel deep learning-based framework for interactive segmentation by incorporating CNNs into a bounding box and scribble-based segmentation pipeline. We propose image-specific fine-tuning to make a CNN model adaptive to a specific test image, which can be either unsupervised (without additional user interactions) or supervised (with additional scribbles). We also propose a weighted loss function considering network and interaction-based uncertainty for the fine-tuning. We applied this framework to two applications: 2D segmentation of multiple organs from fetal MR slices, where only two types of these organs were annotated for training; and 3D segmentation of brain tumor core (excluding edema) and whole brain tumor (including edema) from different MR sequences, where only tumor cores in one MR sequence were annotated for training. Experimental results show that 1) our model is more robust to segment previously unseen objects than state-of-the-art CNNs; 2) image-specific fine-tuning with the proposed weighted loss function significantly improves segmentation accuracy; and 3) our method leads to accurate results with fewer user interactions and less user time than traditional interactive segmentation methods.  

\end{abstract}


%
\IEEEpeerreviewmaketitle

\section{Introduction}
%
%
%
%


\IEEEPARstart{D}{eep} learning with convolutional neural networks (CNNs) has achieved state-of-the-art performance for automated medical image segmentation~\cite{Litjens2017}. However, automatic segmentation methods have not demonstrated sufficiently accurate and robust results for clinical use due to the inherent challenges of medical images, such as poor image quality, 
different imaging and segmentation protocols, and variations among patients~\cite{Zhao2013}. Alternatively, interactive segmentation methods are widely adopted, as integrating the user's knowledge  can take into account the application requirements and make it easier to distinguish different tissues~\cite{Zhao2013, Grady2005, Criminisi2008}. As such, interactive segmentation remains the state of the art for existing commercial surgical planning and navigation products. Though leveraging user interactions often leads to more robust segmentations, a good interactive method should require as little user time as possible to reduce the burden on users. Motivated by these observations, we investigate combining CNNs with user interactions for medical image segmentation to achieve higher segmentation accuracy and robustness with fewer user interactions and less user time. However, there are very few studies on using CNNs for interactive segmentation~\cite{Rajchl2016, Xu2016,Wang2017pami}. This is mainly due to the requirement of large amounts of annotated images for training, the lack of image-specific adaptation and the demanding balance of model complexity, time and memory space efficiency. 
 
The first challenge of using CNNs for interactive segmentation is that current CNNs do not generalize well to previously unseen object classes, as they require labeled instances of each object class to be present in the training set. For medical images, annotations are often expensive to acquire as both expertise and time are needed to produce accurate annotations. This limits the performance of CNNs to segment objects for which annotations are not available in the training stage. 

Second, interactive segmentation often requires image-specific learning to deal with large context variations among different images, but current CNNs are not adaptive to different test images, as 
parameters of the model are learned from training images and then fixed during the testing, without image-specific adaptation.  
It has been shown that image-specific adaptation of a pre-trained Gaussian Mixture Model (GMM) helps to improve segmentation accuracy~\cite{Ribeiro2006}. However, transitioning from simple GMMs to powerful but complex CNNs in this context has not yet been demonstrated.

Third, fast inference and memory efficiency are demanded for interactive methods. These can be relatively easily achieved for 2D segmentations, but become much more problematic for 3D volumes. For example, DeepMedic~\cite{Kamnitsas2017} works on local patches to reduce memory requirements but results in a slow inference. HighRes3DNet~\cite{Li2017} works on an entire volume with relatively fast inference but needs a large amount of GPU memory, leading to high hardware requirements. To make a CNN-based interactive segmentation method efficient to use, enabling CNNs to respond quickly to user interactions and to work on a machine with limited GPU resources (e.g, a standard desktop PC or a laptop) is desirable. DeepIGeoS~\cite{Wang2017pami} combines CNNs with user interactions and has demonstrated good interactivity. However, it has a lack of adaptability to unseen image contexts. 
\begin{figure*}[t]
	\centering
	\includegraphics[width=0.8\linewidth]{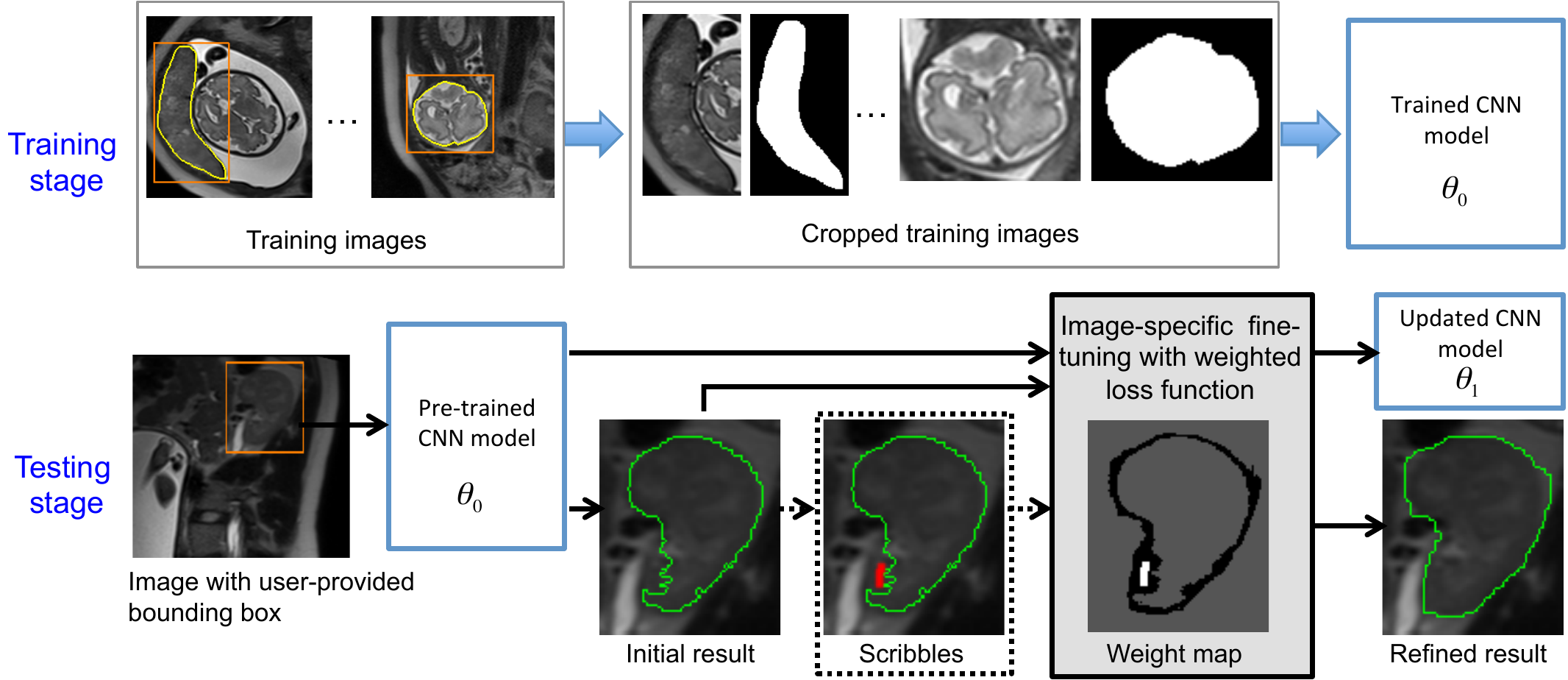}
	\caption[The proposed segmentation method (BIFSeg)]{ 
		The proposed interactive segmentation framework (BIFSeg). 2D images are shown as examples. In the training stage, each instance is cropped with its bounding box, and the CNN model is trained for binary segmentation.
		In the testing stage, image-specific fine-tuning with optional scribbles and a weighted loss function is used. Note that the object class (e.g. a maternal kidney) in the test image may have not been present in the training set. 
	} 
	\label{fig:method}
\end{figure*}
\subsection{Contributions}
 The contributions of this work are four-fold. First, we propose a novel deep learning-based framework for interactive 2D and 3D medical image segmentation by incorporating CNNs into a bounding box and scribble-based binary segmentation pipeline. 
 Second, we propose to use image-specific fine-tuning to adapt a CNN model to each test image independently. 
 The fine-tuning can be either unsupervised (without additional user interactions) or supervised where user-provided scribbles will guide the learning process.  Third, we propose a weighted loss function considering network and interaction-based uncertainty during image-specific fine-tuning. 
 Fourth, we present the first attempt to employ CNNs to segment previously unseen objects. The proposed framework does not require annotations of all the organs for training. Thus, it can be applied to new organs or new segmentation protocols directly.
 

\subsection{Related Works}
\subsubsection{CNNs for Image Segmentation}
For natural image segmentation, FCN~\cite{Long2014} and DeepLab~\cite{Chen2015iclr} are among the state-of-the-art performing methods. For 2D biomedical image segmentation, efficient networks such as U-Net~\cite{Hefny2015a}, DCAN~\cite{chen_hao2016_dcan} and Nabla-net~\cite{RichardMckinley2016} have been proposed. For 3D volumes, patch-based CNNs were proposed for segmentation of the brain tumor~\cite{Kamnitsas2017} and pancreas~\cite{Roth2015}, 
and more powerful end-to-end 3D CNNs were proposed by V-Net~\cite{Milletari2016}, HighRes3DNet~\cite{Li2017}, and 3D deeply supervised network~\cite{Dou2017}.

\subsubsection{Interactive Segmentation Methods}
An extensive range of interactive segmentation methods have been proposed~\cite{Zhao2013}. Representative methods include Graph Cuts~\cite{Boykov2001}, 
Random Walks~\cite{Grady2005} and GeoS~\cite{Criminisi2008}. Machine learning methods have been widely used to achieve high accuracy and interaction efficiency. For example, GMMs are used by GrabCut~\cite{Rother2004} to segment color images. Online random forests (ORFs) are employed by SlicSeg~\cite{Wang2016} for segmentation of fetal MRI volumes. In~\cite{Top2011a}, active learning is used to segment 3D Computed Tomography (CT) images. They have achieved more accurate segmentations with fewer user interactions compared with traditional interactive segmentation methods. 

To combine user interactions with CNNs, DeepCut~\cite{Rajchl2016} and ScribbleSup~\cite{Lin2016} propose to leverage user-provided bounding boxes or scribbles, but they employ user interactions as sparse annotations for the training set rather than as guidance for dealing with a single test image. 3D U-Net~\cite{Abdulkadir2016} learns from annotations of some slices in a volume and produces a dense 3D segmentation, but takes a long time for training and cannot be made responsive to user interactions.
In~\cite{Xu2016}, an FCN is combined with user interactions for 2D RGB image segmentation, without adaptation for medical images. DeepIGeoS~\cite{Wang2017pami} uses geodesic distance transforms of scribbles as additional channels of CNNs for interactive medical image segmentation, but cannot deal with previously unseen object classes.

\subsubsection{Model Adaptation}
Previous learning-based interactive segmentation methods often employ an image-specific model. For example, GrabCut~\cite{Rother2004} and SlicSeg~\cite{Wang2016} learn from the target image with GMMs and ORFs, respectively, so that they can be well adapted to the specific target image. Learning a model from a training set with image-specific adaptation during the testing has also been used to improve the segmentation performance. For example, an adaptive GMM has been used to address the distribution mismatch between a test image and the training set~\cite{Ribeiro2006}. For CNNs, fine-tuning~\cite{Tajbakhsh2016} is used for domain-wise model adaptation to address the distribution mismatch between different training sets. However, to the best of our knowledge, this paper is the first work to propose image-specific model adaptation for CNNs.
\begin{figure*}[t]
	\centering
\includegraphics[width=0.9\linewidth]{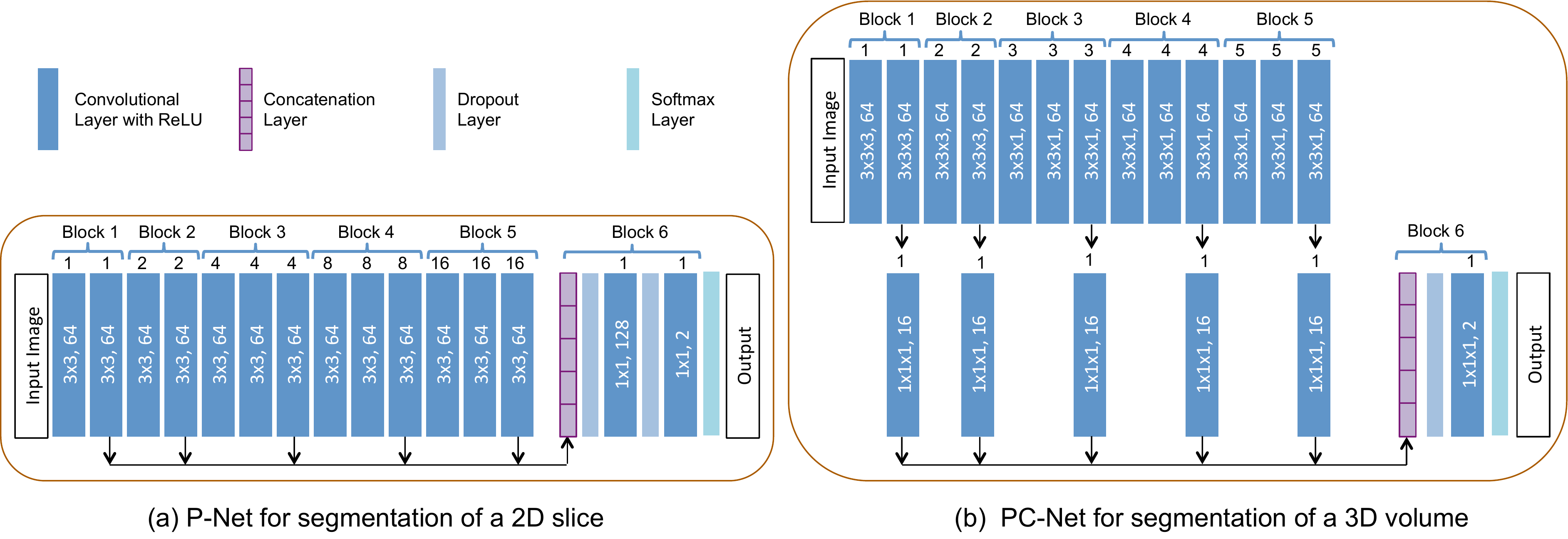}	\caption[Resolution-preserving network]{ 
		Our resolution-preserving networks with dilated convolution for 2D segmentation (a) and 3D segmentation (b). The numbers in each dark blue box denote convolutional kernel size and number of output channels, and the number on the top denotes dilation parameter.
	} 
	\label{fig:network}
\end{figure*}
\section{Method}
The proposed interactive segmentation framework 
is depicted in Fig.~\ref{fig:method}. We refer to it as BIFSeg. To deal with different (including previously unseen) objects in a unified framework, we propose to use a CNN that takes as input the content of a bounding box of one instance and gives a binary segmentation. During the test stage, the bounding box is provided by the user, and the segmentation and the CNN are alternatively refined through unsupervised (without additional user interactions) or supervised (with user-provided scribbles) image-specific fine-tuning. Our framework is general, flexible and can handle both 2D and 3D segmentations with few assumptions of network structures. In this paper, we choose to use the state-of-the-art network structures proposed in~\cite{Wang2017pami}. The contribution of BIFSeg is nonetheless largely different from~\cite{Wang2017pami} as BIFSeg focuses on segmentation of previously unseen object classes and fine-tunes the CNN model on the fly for image-wise adaptation that can be guided by user interactions.
\subsection{CNN Models}

For 2D images, we adopt the P-Net~\cite{Wang2017pami} for bounding box-based binary segmentation. The network is  resolution-preserving using dilated convolution~\cite{Chen2015iclr} to avoid potential loss of details. 
As shown in Fig.~\ref{fig:network}(a), it consists of six blocks with a receptive field of 181$\times$181. The first five blocks have dilation parameters of 1, 2, 4, 8 and 16, respectively, so they capture features at different scales. Features from these five blocks are concatenated and fed into block6 that serves as a classifier. A softmax layer is used to obtain probability-like outputs.
In the testing stage, we update the model based on image-specific fine-tuning. To ensure efficient fine-tuning and fast response to user interactions, we only fine-tune parameters of the classifier (block6). Thus, features in the concatenation layer for the test image can be stored before the fine-tuning.

For 3D images, we consider a trade-off between receptive field, inference time and memory efficiency. As shown in Fig.~\ref{fig:network}(b), the network is similar to P-Net. It has an anisotropic receptive field 85$\times$85$\times$9. Compared with slice-based networks, it employs 3D context. Compared with large isotropic 3D receptive fields~\cite{Li2017}, it has less memory consumption during inference~\cite{Wang17brats}. Besides, anisotropic acquisition is often used in MR images. We use 3$\times$3$\times $3 kernels in the first two blocks and 3$\times$3$\times $1 kernels in block3 to block5. Similar to P-Net, we fine-tune the classifier (block6) with pre-computed concatenated features. To save space for storing the concatenated features, we use 1$\times$1$\times$1 convolutions to compress the features in block1 to block5 and then concatenate them.  We refer to this 3D network with feature compression as PC-Net.

\subsection{Training of CNNs}
The training stage for 2D/3D segmentation is shown in the first row of Fig.~\ref{fig:method}.
Consider a $K$-ary segmentation training set $T=\{(X_1,Y_1), (X_2, Y_2), ...\}$ where $X_p$ is one training image and $Y_p$ is the corresponding label map. The label set of $T$ is $\{0, 1, 2, ..., K-1\}$ with 0 being the background label. Let $N_k$ denote the number of instances of the $k$th object type, so the total number of instances is $\hat{N} = \sum_k{N_k}$. Each image $X_p$ can have instances of multiple object classes. Suppose the label of the $q$th instance in $X_p$ is $l_{pq}$, $Y_p$ is converted into a binary image $Y_{pq}$ based on whether the value of each pixel in  $Y_p$ equals to $l_{pq}$. The bounding box $B_{pq}$ of that training instance is automatically calculated based on $Y_{pq}$ and expanded by a random margin in the range of 0 to 10 pixels/voxels. $X_p$ and $Y_{pq}$ are cropped based on $B_{pq}$.
Thus, $T$ is converted into a cropped set $\hat{T}=\{(\hat{X}_1,\hat{Y}_1), (\hat{X}_2, \hat{Y}_2), ...\}$ with size $\hat{N}$ and label set $\{0, 1\}$ where 1 is the label of the instance foreground and 0 the background. With $\hat{T}$, the CNN model (e.g, P-Net or PC-Net) is trained to extract the target from its bounding box, which is a binary segmentation problem irrespective of the object type. A cross entropy loss function is used for training.

\subsection{Unsupervised and Supervised Image-specific Fine-tuning}
In the testing stage, let $\hat{X}$ denote the sub-image inside a user-provided bounding box and $\hat{Y}$ be the target label of $\hat{X}$. The set of parameters of the trained CNN is $\theta$. With the initial segmentation $\hat{Y}_0$ obtained by the trained CNN, the user may provide (i.e., supervised) or not provide (i.e., unsupervised) a set of scribbles to guide the update of $\hat{Y}_0$. Let $S^f$ and $S^b$ denote the scribbles for foreground and background, respectively, so the entire set of scribbles is $S = S^f \cup S^b$. Let $s_i$ denote the user-provided label of a pixel in the scribbles, then we have $s_i=1$ if $i\in S_f$ and $s_i=0$ if $i\in S_b$. We minimize an objective function that is similar to GrabCut~\cite{Rother2004} but we use P-Net or PC-Net instead of a GMM:
\begin{equation}
\begin{aligned}
& \underset{\hat{Y}, \theta}{\text{arg min}} \left\{	E(\hat{Y},\theta) = \sum_{i}\phi(\hat{y}_i|\hat{X},\theta) + \lambda\sum_{i,j}\psi(\hat{y}_i, \hat{y}_j|\hat{X})\right\}
	\\
&\text{subject to}: \hat{y}_i = s_i \quad \text{if } i \in S 
	\label{eq:crf_energy} 
\end{aligned}
\end{equation}
where $E(\hat{Y},\theta)$ is constrained by user interactions if $S$ is not empty. $\phi$ and $\psi$ are the unary and pairwise energy terms, respectively.
$\lambda$ is the weight of $\psi$. 
An unconstrained optimization of an energy similar to $E$ is used in~\cite{Rajchl2016} for weakly supervised learning. In that work, the energy was based on the probability and label map of all the images in a training set, which is a different task from ours, as we focus on a single test image.
We follow a typical choice of $\psi$~\cite{Boykov2001}:
\begin{align}
	\psi(\hat{y}_i, \hat{y}_j|\hat{X})= [\hat{y}_i \neq \hat{y}_j] \text{exp}\left(-\frac{(\hat{X}(i)-\hat{X}(j))^2}{2\sigma^2}\right)\cdot \frac{1}{d_{ij}}
	\label{eq:pairwise_potential}
\end{align}
where $[\cdot]$ is 1 if $\hat{y}_i \neq \hat{y}_j$ and 0 otherwise. $d_{ij}$ is the Euclidean distance between pixel $i$ and pixel $j$. $\sigma$ controls the effect of intensity difference. $\phi$ is defined as:
\begin{align}
\phi(\hat{y}_i|\hat{X},\theta)=
-\text{log}P(\hat{y}_i|\hat{X},\theta)  
\label{eq:unary_potential}
\end{align}
where 
$P(\hat{y}_i|\hat{X},\theta)$ is the probability given by softmax output of the CNN. Let $p_i=P(\hat{y}_i = 1|\hat{X},\theta)$ be the probability of pixel $i$ belonging to the foreground, we then have:
\begin{equation}
\begin{aligned}
\text{log} P(\hat{y}_i|\hat{X},\theta) = \hat{y}_i\text{log} p_i +  
(1-\hat{y}_i)\text{log} (1 - p_i)
\label{eq:cross_entropy}
\end{aligned}
\end{equation}

The optimization of Eq.~\eqref{eq:crf_energy} can be decomposed into steps that alternatively update the segmentation label $\hat{Y}$ and network parameters $\theta$~\cite{Rajchl2016,Rother2004}. In the label update step, we fix $\theta$ and solve for $\hat{Y}$, and Eq.~\eqref{eq:crf_energy} becomes a CRF problem:  
\begin{equation}
\begin{aligned}
& \underset{\hat{Y}}{\text{arg min}} \left\{	E(\theta) = \sum_{i}\phi(\hat{y}_i|\hat{X},\theta) + \lambda\sum_{i,j}\psi(\hat{y}_i, \hat{y}_j|\hat{X})\right\}
\\
&\text{subject to}: \hat{y}_i = s_i \quad \text{if } i \in S 
\label{eq:label_step_constrainted} 
\end{aligned}
\end{equation}
For implementation ease, the constrained optimization in Eq.~\eqref{eq:label_step_constrainted} is converted to an unconstrained equivalent: 
\begin{equation}
\begin{aligned}
 \underset{\hat{Y}}{\text{arg min}} \left\{ \sum_{i}\phi'(\hat{y}_i|\hat{X},\theta) + \lambda\sum_{i,j}\psi(\hat{y}_i, \hat{y}_j|\hat{X}) \right\}
\label{eq:label_step} 
\end{aligned}
\end{equation}
\begin{align}
\phi'(\hat{y}_i|\hat{X},\theta)=
\begin{cases}
+\infty & \quad \text{if } i \in S \text{ and }\hat{y}_i=s_i\\
0   & \quad \text{if } i \in S \text{ and }\hat{y}_i\neq s_i\\
-\text{log} P(\hat{y}_i|\hat{X},\theta)  & \quad \text{otherwise}\\
\end{cases}
\label{eq:unary_potential2}
\end{align}
Since $\theta$ and therefore $\phi'$ are fixed, and $\psi$ is submodular, Eq.~\eqref{eq:label_step} can be solved by Graph Cuts~\cite{Boykov2001}. 
In the network update step, we fix $\hat{Y}$ and solve for $\theta$:
\begin{equation}
\begin{aligned}
 &\underset{\theta}{\text{arg min}}\left\{ E(\hat{Y})=\sum_{i}\phi(\hat{y}_i|\hat{X},\theta) \right\}\\
&\text{subject to}: \hat{y}_i = s_i \quad \text{if } i \in S 
\label{eq:network_step} 
\end{aligned}
\end{equation}
Thanks to the constrained optimization in Eq.~\eqref{eq:label_step_constrainted}, the label update step necessarily leads to $\hat{y}_i = s_i$ for $i\in S$. Eq. \eqref{eq:network_step} can be treated as an unconstrained optimization:
\begin{equation}
\begin{aligned}
\underset{\theta}{\text{arg min}} \left\{- \sum_{i}\Big(\hat{y}_i\text{log} p_i + (1-\hat{y}_i)\text{log}(1-p_i)\Big)\right\}
\label{eq:cross_entropy_loss} 
\end{aligned}
\end{equation}

\subsection{Weighted Loss Function during Network Update Step}\label{sec:weighted_loss}
During the network update step, the CNN is fine-tuned to fit the current segmentation $\hat{Y}$. Compared with a standard learning process that treats all the pixels equally, we propose to weight different kind of pixels considering their confidence. First, user-provided scribbles have much higher confidence than the other pixels, and they should have a higher impact on the loss function, leading to a weighted version of Eq.~\eqref{eq:unary_potential}:
\begin{align}
\phi(\hat{y}_i| \hat{X},\theta) = -w(i)\text{log}P(\hat{y}_i|\hat{X},\theta)
\label{eq:new_unary_term}
\end{align}
\begin{align}
w(i) = 
\begin{cases}
\omega       & \quad \text{if } i\in S \\
1  & \quad \text{otherwise}\\
\end{cases}
\label{eq:weight_definition0}
\end{align}
where $\omega \geq 1$ is the weight associated with scribbles. $\phi$ defined in Eq.~\eqref{eq:new_unary_term} allows Eq.~\eqref{eq:label_step_constrainted} to remain unchanged for the label update step. In the network update step, Eq.~\eqref{eq:cross_entropy_loss} becomes:
\begin{equation}
\begin{aligned}
\underset{\theta}{\text{arg min}} \left\{- \sum_{i}w(i)\Big(\hat{y}_i\text{log} p_i + (1-\hat{y}_i)\text{log}(1-p_i)\Big)\right\}
\label{eq:weighted_loss} 
\end{aligned}
\end{equation}
Note that the energy optimization problem of Eq.~\eqref{eq:crf_energy} remains well-posed with Eq.~\eqref{eq:new_unary_term},~\eqref{eq:weight_definition0}, and~\eqref{eq:weighted_loss}. 

Second, $\hat{Y}$ may contain mis-classified pixels that can mis-lead the network update process. To address this problem, we propose to fine-tune the network by ignoring pixels with high uncertainty (low confidence) in the test image. We propose to use network-based uncertainty and scribble-based uncertainty. The network-based uncertainty is based on the network's softmax output. 
Since $\hat{y}_i$ is highly uncertain (has low confidence) if $p_i$ is close to 0.5, we define the set of pixels with high network-based uncertainty as $U_p = \{i|t_0 < p_i < t_1\}$ where $t_0$ and $t_1$ are the lower and higher threshold values of foreground probability, respectively. The scribble-based uncertainty is based on the geodesic distance to scribbles.
Let $G(i, S^f)$ and $G(i, S^b)$ denote the geodesic distance~\cite{Criminisi2008} from pixel $i$ to $S^f$ and $S^b$, respectively. Since the scribbles are drawn on mis-segmented areas for refinement, it is likely that pixels close to $S$ have been incorrectly labeled by the initial segmentation. Let $\epsilon$ be a threshold value for the geodesic distance. We define the set of pixels with high scribble-based uncertainty as $U_s = U_s^f \cup U_s^b$ where $U_s^f = \{i|i\notin S, G(i, S^f)<\epsilon, \hat{y_i}=0\}$, $U_s^b = \{i|i\notin S, G(i, S^b)<\epsilon, \hat{y_i}=1\}$. Therefore, a full version of the weighting function is (an example is shown in Fig.~\ref{fig:weight_map}):
\begin{figure}[t]
	\centering
\includegraphics[width=1.0\linewidth]{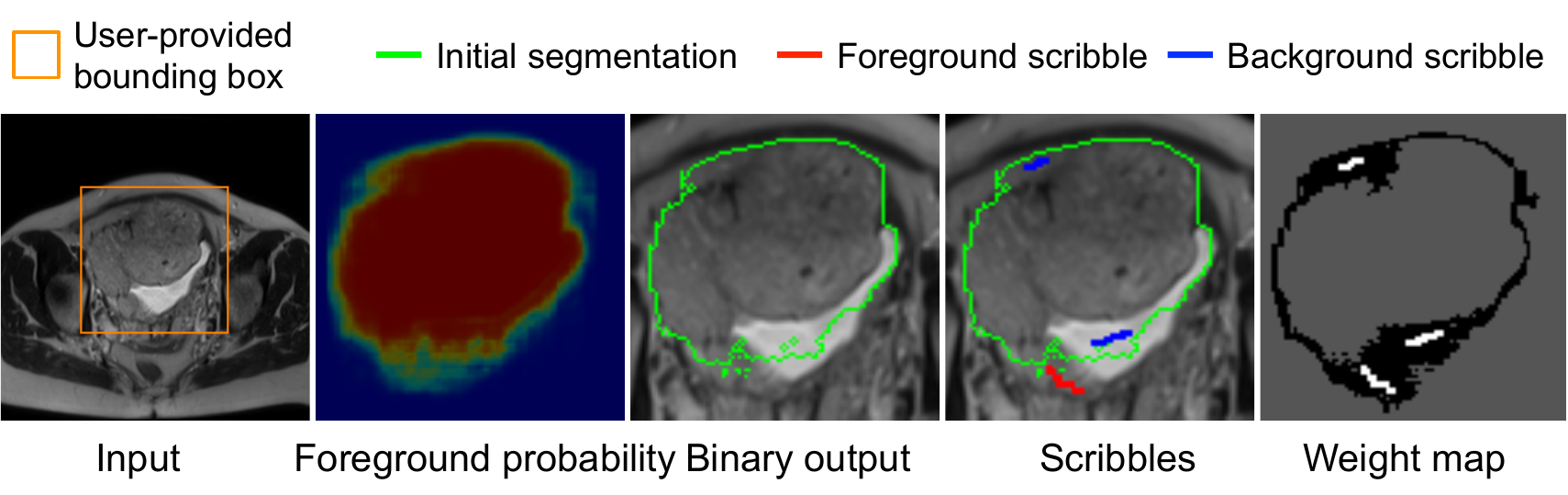}
	\caption[An example of weighting function for fine-tuning]{ 
		An example of weight map for image-specific fine-tuning. The weight is 0 for pixels with high uncertainty (black), $\omega$ for scribbles (white), and 1 for the remaining pixels (gray).
	} 
	\label{fig:weight_map}
\end{figure}
\begin{align}
	w(i) =   
	\begin{cases}
		\omega       & \quad \text{if } i\in S \\
		0       & \quad \text{if } i\in U_p \cup U_s \\
		1  & \quad \text{otherwise}\\
	\end{cases}
	\label{eq:weight_definition}
\end{align}
The new definition of $w(i)$ is well motivated in the network update step. However, in the label update step, introducing zero unary weights in Eq.~\eqref{eq:label_step_constrainted} would make the label update of corresponding pixels entirely driven by the pairwise potentials. Therefore, we choose to keep Eq.~\eqref{eq:label_step_constrainted} unchanged. 
\subsection{Implementation Details}
We used the Caffe\footnote{http://caffe.berkeleyvision.org}~\cite{Jia2014} library to implement our P-Net and PC-Net. The training process was done via one node of the Emerald cluster\footnote{http://www.ses.ac.uk/high-performance-computing/emerald} with two 8-core E5-2623v3 Intel Haswells, a K80 NVIDIA GPU and 128GB memory. Stochastic gradient decent was used for training,  with momentum 0.9, batch size 1, weight decay $5\times10^{-4}$, maximal number of iterations 60k, initial learning $10^{-3}$ that was halved every 5k iterations. For each application, the images in each modality were normalized by the mean value and standard variation of the training images. During training, the bounding box for each object was automatically generated based on the ground truth label with a random margin in the range of 0 to 10 pixels/voxels.

For the testing with user interactions, the trained CNN models were deployed to a MacBook Pro (OS X 10.9.5) with 16GB RAM, an Intel Core i7 CPU running at 2.5GHz and an NVIDIA GeForce GT 750M GPU. A Matlab GUI and a PyQt GUI were used for user interactions on 2D and 3D images, respectively. The bounding box  was provided by the user. For image-specific fine-tuning, $\hat{Y}$ and $\theta$ were alternatively updated for four iterations. In each network update step, we used a learning rate $10^{-2}$ and iteration number 20. We used a grid search with the training data to get proper values of $\lambda$, $\sigma$, $t_0$, $t_1$, $\epsilon$ and $\omega$. Their numerical values are listed in the specific experiments sections \ref{sec:2d_segmentation} and \ref{sec:3d_segmentation}.

\section{Experiments and Results}

\begin{figure}[t]
	\centering
\includegraphics[width=1.0\linewidth]{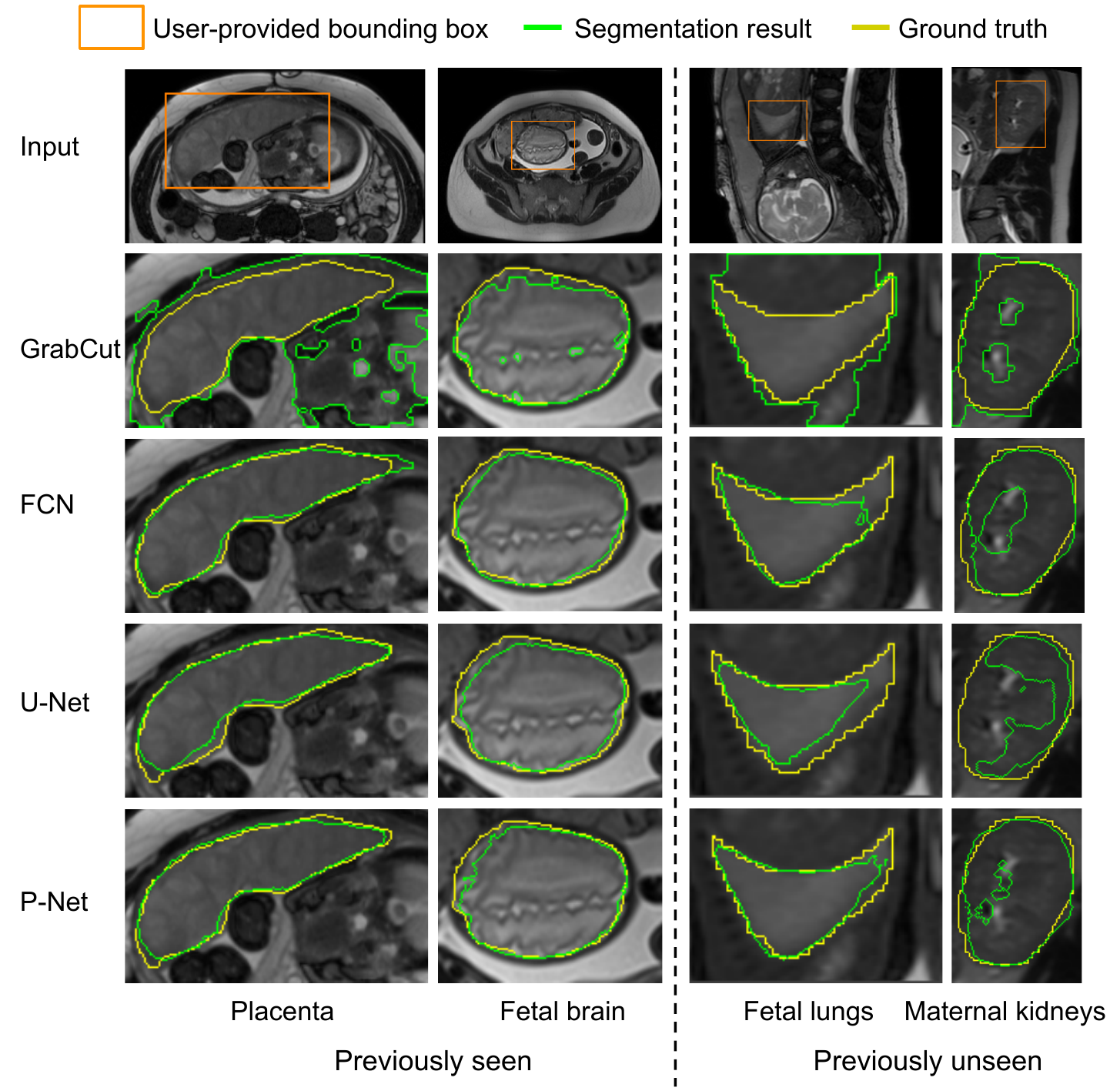}
	\caption[Visual comparison between BIFSeg and counterparts for segmentation using bounding box]{ 
		Visual comparison of initial segmentation of multiple organs from fetal MRI with a bounding box. 
		All the methods use the same bounding box for each test instance. Note that fetal lungs and maternal kidneys are previously unseen objects but P-Net works well on them.
	} 
	\label{fig:fetal_different_net}
\end{figure}

\begin{table}
	\centering
	\footnotesize
	\caption{Quantitative comparison of initial fetal MRI segmentation from a bounding box. $T_m$ is the machine time. $\wedge$ denotes previously unseen objects. In each row, bold font denotes the best value. * denotes $p$-value $<$ 0.05 compared with the others.}
	\label{tab:fetal_different_net}
	\begin{tabular}{p{0.4cm}|p{0.4cm}|p{1.2cm}|p{1.4cm}|p{1.4cm}|p{1.4cm}}
		\hline
		\multicolumn{2}{l|}{} & FCN & U-Net  & P-Net  & GrabCut \\ \hline
		\multirow{4}{*}{\vtop{\hbox{\strut Dice}\hbox{\strut (\%)}}} & P & \bf{85.31$\pm$8.73} & 82.86$\pm$9.85  & 84.57$\pm$8.37 & 62.90$\pm$12.79 \\ 
		& FB & \bf{89.53$\pm$3.91} & 89.19$\pm$5.09 & 89.44$\pm$6.45 & 83.86$\pm$14.33 \\
	    & FL$^{\wedge}$ & 81.68$\pm$5.95 & 80.64$\pm$6.10 & \bf{83.59$\pm$6.42*} & 63.99$\pm$15.86 \\
		& MK$^{\wedge}$ & 83.58$\pm$5.48 & 75.20$\pm$11.23 & \bf{85.29$\pm$5.08*} &  73.85$\pm$7.77\\  \hline 
		\multicolumn{2}{c|}{ $T_m$(s)} & \bf{0.11$\pm$0.04*} & 0.24$\pm$0.07 & 0.16$\pm$0.05 &  1.62$\pm$0.42\\  \hline 
		\multicolumn{6}{@{}l}{P: Placenta, FB: Fetal brain, FL: Fetal lungs, MK: Maternal kidneys.} 
	\end{tabular}
	
\end{table}
\begin{figure}[t]
	\centering
\includegraphics[width=1.0\linewidth]{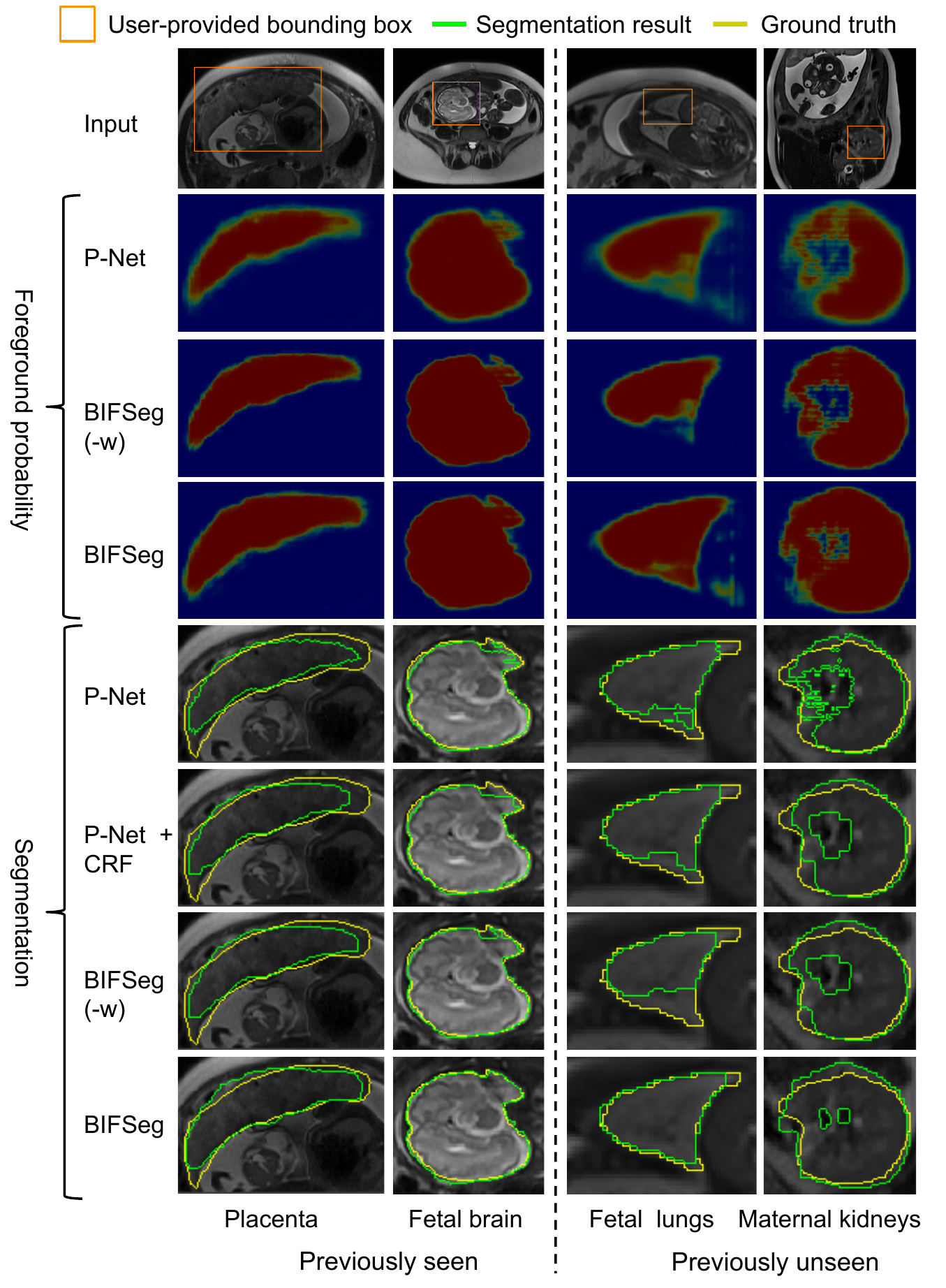}
	\caption[Visual comparison between BIFSeg and counterparts for segmentation using bounding box]{ 
		Visual comparison of P-Net and three unsupervised refinement methods for fetal MRI segmentation. The foreground probability is visualized by heatmap.
	} 
	\label{fig:fetal_unsupervised_seg}
\end{figure}

We validated the proposed framework with two applications: 2D segmentation of multiple organs from fetal MRI and 3D segmentation of brain tumors from contrast enhanced T1-weighted (T1c) and Fluid-attenuated Inversion Recovery (FLAIR) images. For both applications,   
we additionally investigated the segmentation performance on previously unseen objects that were not present in the training set. 

\subsection{Comparison Methods and Evaluation Metrics}
To investigate the performance of different networks with the same bounding box, we compared P-Net with FCN~\cite{Long2014} and U-Net~\cite{Hefny2015a} for 2D images, and compared PC-Net with DeepMedic~\cite{Kamnitsas2017} and HighRes3DNet~\cite{Li2017} for 3D images\footnote{DeepMedic and HighRes3DNet were implemented in http://niftynet.io}. The original DeepMedic works on multiple modalities, and we adapted it to work on a single modality. All these methods were evaluated on the laptop during the testing except for HighRes3DNet that was run on the cluster due to the laptop's limited GPU memory. To validate the proposed unsupervised/supervised image-specific fine-tuning, we compared BIFSeg with 1) the initial output of P-Net/PC-Net, 2) post-processing the initial output with a CRF (using user interactions as hard constraints if they were given), and 3) image-specific fine-tuning based on Eq.~\eqref{eq:crf_energy} with $w(i) =1 $ for all the pixels, which is referred to as BIFSeg(-w). 

BIFSeg was also compared with other interactive segmentation methods: GrabCut~\cite{Rother2004}, SlicSeg~\cite{Wang2016} and Random Walks~\cite{Grady2005} for 2D segmentation, and GeoS~\cite{Criminisi2008}, GrowCut~\cite{Vezhnevets2005} and 3D GrabCut~\cite{Ram2013} for 3D segmentation. 
The 2D/3D GrabCut used the same bounding box as used by BIFSeg, and they used 3 and 5 components for the foreground and background GMMs, respectively. 
SlicSeg, Random Walks, GeoS and GrowCut require scribbles without a bounding box for segmentation. The segmentation results by an Obstetrician and a Radiologist 
were used for evaluation. For each method, each user provided scribbles to update the result multiple times until the user accepted it as the final segmentation.
The Dice score between a segmentation and the ground truth was used for quantitative evaluations:
$\text{Dice}=2|\mathcal{R}_a\cap \mathcal{R}_b|/(|\mathcal{R}_a|+|\mathcal{R}_b|)
$
where $\mathcal{R}_a$ and $\mathcal{R}_b$ denote the region segmented by an algorithm and the ground truth, respectively. The $p$-value between different methods was computed by the Student's $t$-test.

\subsection{2D Segmentation of Multiple Organs from Fetal MRI}\label{sec:2d_segmentation}
\subsubsection{Data}
Single-shot Fast Spin Echo (SSFSE) was used to acquire stacks of T2-weighted MR images from 18 pregnant women with pixel size 0.74 to 1.58 mm and inter-slice spacing 3 to 4 mm. Due to the large inter-slice spacing and inter-slice motion, interactive 2D segmentation is more suitable than direct 3D segmentation~\cite{Wang2016}. The placenta and fetal brain from 10 volumes (356 slices) were used for training. The other 8 volumes (318 slices) were used for testing. From the test images, we aimed to segment the placenta, fetal brain, and previously unseen fetal lungs and maternal kidneys. 
Manual segmentations by a Radiologist were used as the ground truth. P-Net was used for this segmentation task. To deal with organs at different scales, we resized the input of P-Net so that the minimal value of width and height was 128 pixels. Parameter setting was $\lambda$
= 3.0, $\sigma$ = 0.1, $t_0$ = 0.2, $t_1$ = 0.7, $\epsilon$ = 0.2, $\omega$ = 5.0 based on a grid search with the training data. 

\subsubsection{Initial Segmentation based on P-Net}
Fig.~\ref{fig:fetal_different_net} shows the initial segmentation of different organs from fetal MRI with user-provided bounding boxes. 
It can be observed that GrabCut achieves a poor segmentation except for the fetal brain where there is a good contrast between the target and the background. For the placenta and fetal brain, FCN, U-Net and P-Net achieves visually similar results that are close to the ground truth. However, for fetal lungs and maternal kidneys that are previously unseen in the training set, FCN and U-Net lead to a large region of under-segmentation. In contrast, P-Net performs noticeably better than FCN and U-Net when dealing with these two unseen objects. A quantitative evaluation of these methods are listed in Table~\ref{tab:fetal_different_net}. It shows that P-Net achieves the best accuracy for unseen fetal lungs and maternal kidneys with average machine time 0.16s.

\subsubsection{Unsupervised Image-specific Fine-tuning}

For unsupervised refinement, the initial segmentation result obtained by P-Net was refined by CRF, BIFSeg(-w) and BIFSeg without additional scribbles, respectively. The results are shown in Fig.~\ref{fig:fetal_unsupervised_seg}. The second to fourth rows show the foreground probability obtained by P-Net before and after the fine-tuning. In the second row, the initial output of P-Net has a probability around 0.5 for many pixels, which indicates a high uncertainty. After image-specific fine-tuning, most pixels in the outputs of BIFSeg(-w) and BIFSeg have a probability close to 0.0 or 1.0.
The remaining rows show the segmentations by P-Net and the three refinement methods, respectively. 
The visual comparison shows that BIFSeg performs better than P-Net + CRF and BIFSeg(-w). Quantitative measurements are presented in Table~\ref{tab:fetal_unsupervised_seg}. It shows that BIFSeg achieves a larger improvement of accuracy from the initial segmentation when compared with the use of CRF or BIFSeg(-w). In this 2D case, BIFSeg takes 0.72s in average for unsupervised image-specific fine-tuning.
\begin{figure}[t]
	\centering
\includegraphics[width=1.0\linewidth]{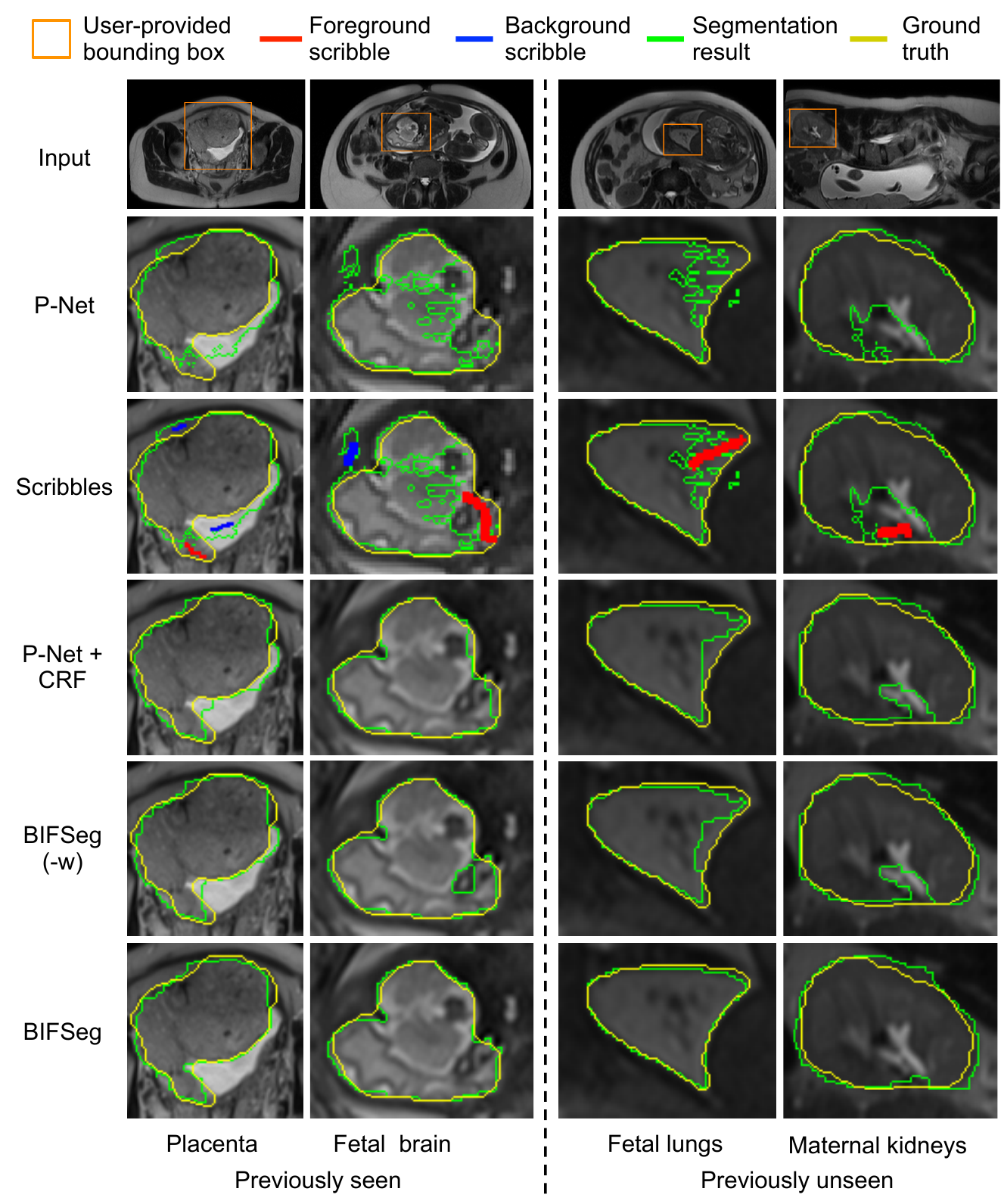}
	\caption[Comparison between different segmentation methods using user-provided bounding box with scribbles]{ 
		Visual comparison of P-Net and three supervised refinement methods for fetal MRI segmentation. The same initial segmentation and scribbles are used for P-Net + CRF, BIFSeg(-w) and BIFSeg. 
	} 
	\label{fig:fetal_supervised_seg}
\end{figure}

\begin{figure}[t]
	\centering 
	\centering
\includegraphics[width=1.0\linewidth]{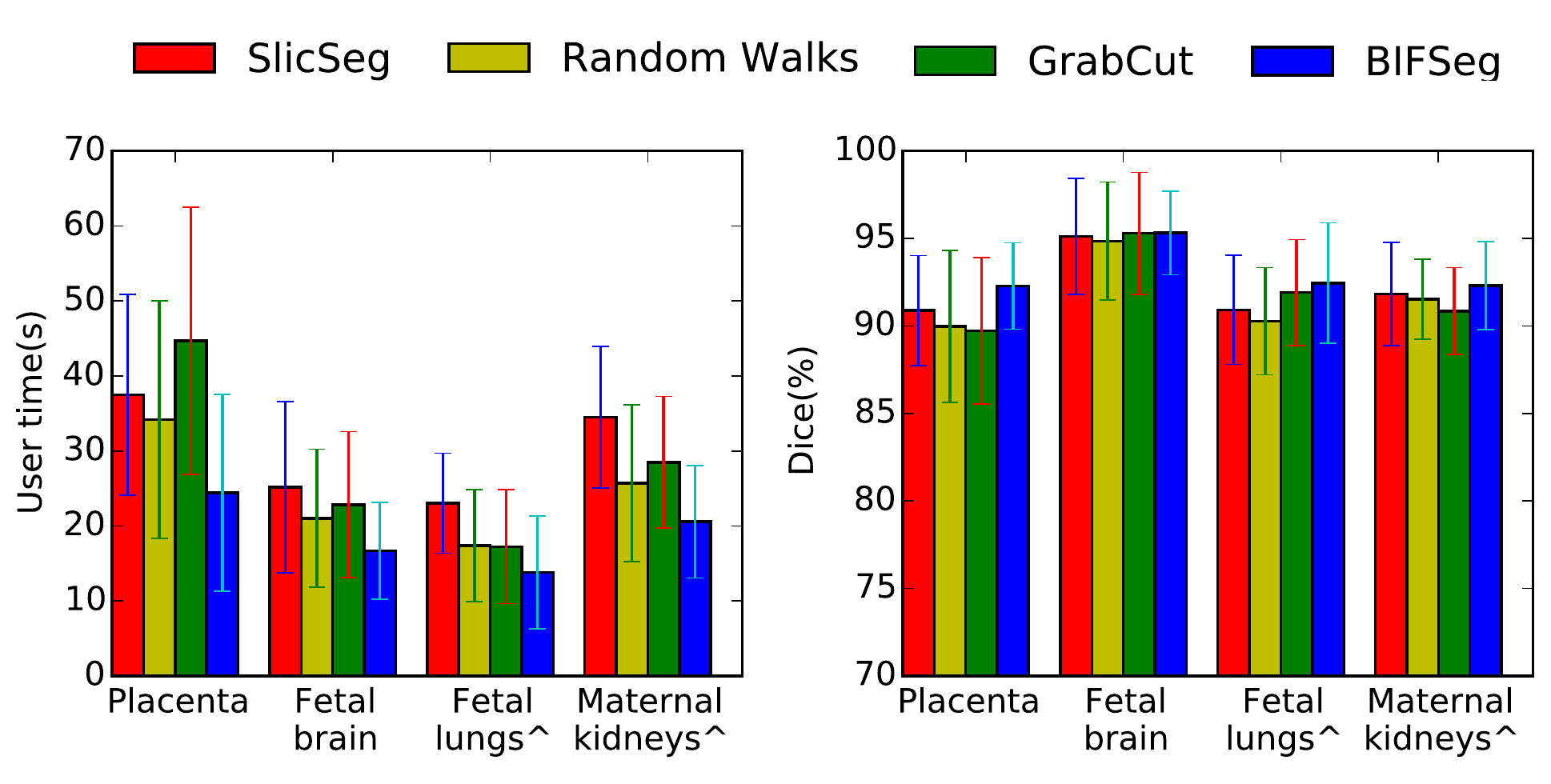}
	\caption[Comparison between different segmentation methods using user-provided bounding box with scribbles]{ 
		User time and Dice score of different interactive methods for fetal MRI segmentation. $^{\wedge}$ denotes previously unseen objects for BIFSeg.
	} 
	\label{fig:fetal_compare_grabcut}
\end{figure}

\begin{table}
	\centering
	\footnotesize
	\caption{Quantitative comparison of P-Net and three unsupervised refinement methods for fetal MRI segmentation. $T_m$ is the machine time for refinement. $\wedge$ denotes previously unseen objects. In each row, bold font denotes the best value. * denotes $p$-value $<$ 0.05 compared with the others. }
	\label{tab:fetal_unsupervised_seg}
	\begin{tabular}{@{}p{0.4cm}|p{0.4cm}|p{1.2cm}|p{1.4cm}|p{1.4cm}|p{1.4cm}}
		\hline
		\multicolumn{2}{l|}{} &  P-Net & P-Net+CRF & BIFSeg(-w) & BIFSeg \\ \hline
		\multirow{4}{*}{\vtop{\hbox{\strut Dice}\hbox{\strut (\%)}}}
		& P & 84.57$\pm$8.37 & 84.87$\pm$8.14& 82.74$\pm$10.91 & \bf{86.41$\pm$7.50*}\\ 
		& FB  & 89.44$\pm$6.45 & 89.55$\pm$6.52 & 89.09$\pm$8.08 & \bf{90.39$\pm$6.44}\\
		& FL$^{\wedge}$ & 83.59$\pm$6.42 & 83.87$\pm$6.52 & 82.17$\pm$8.87 & \bf{85.35$\pm$5.88*}\\
		& MK$^{\wedge}$ & 85.29$\pm$5.08 & 85.45$\pm$5.21 & 84.61$\pm$6.21& \bf{86.33$\pm$4.28*}\\  
		\hline
		\multicolumn{2}{c|}{$T_m$ (s)} & - & \bf{0.02$\pm$0.01*} & 0.71$\pm$0.12 &  0.72$\pm$0.12\\  \hline 
		\multicolumn{6}{@{}l}{P: Placenta, FB: Fetal brain, FL: Fetal lungs, MK: Maternal kidneys.} 
	\end{tabular}
\end{table}

\begin{table}
	\centering
	\footnotesize
	\caption{Quantitative comparison of P-Net and three supervised refinement methods with scribbles for fetal MRI segmentation. $T_m$ is the machine time for refinement. $\wedge$ denotes previously unseen objects. In each row, bold font denotes the best value. * denotes $p$-value $<$ 0.05 compared with the others. }
	\label{tab:fetal_supervised_seg}
	\begin{tabular}{p{0.4cm}|p{0.4cm}|p{1.2cm}|p{1.4cm}|p{1.4cm}|p{1.4cm}}
		\hline
		\multicolumn{2}{l|}{}  & P-Net & P-Net+CRF & BIFSeg(-w) & BIFSeg \\ \hline
		\multirow{4}{*}{\vtop{\hbox{\strut Dice}\hbox{\strut (\%)}}}
		& P & 84.57$\pm$8.37 & 88.64$\pm$5.84& 89.79$\pm$4.60 & \bf{91.93$\pm$2.79*}\\ 
		& FB & 89.44$\pm$6.45 & 94.04$\pm$4.72 & 95.31$\pm$3.39 & \bf{95.58$\pm$1.94}\\
		& FL$^{\wedge}$ & 83.59$\pm$6.42 & 88.92$\pm$3.87 & 89.21$\pm$2.95 & \bf{91.71$\pm$3.18*}\\
		& MK$^{\wedge}$ & 85.29$\pm$5.08 & 87.51$\pm$4.53 & 87.78$\pm$4.46& \bf{89.37$\pm$2.31*}\\  
		\hline
		\multicolumn{2}{c|}{$T_m$ (s)} & - & \bf{0.02$\pm$0.01*} & 0.72$\pm$0.11 &  0.74$\pm$0.12\\  \hline 
		\multicolumn{6}{@{}l}{P: Placenta, FB: Fetal brain, FL: Fetal lungs, MK: Maternal kidneys.} 
	\end{tabular}
\end{table}
\subsubsection{Supervised Image-specific Fine-tuning} 
Fig.~\ref{fig:fetal_supervised_seg} shows examples of supervised refinement with additional scribbles. The second row shows the initial segmentation obtained by P-Net. In the third row, red and blue scribbles are drawn in mis-segmented regions to label the corresponding pixels as the foreground and background, respectively. 
The same initial segmentation and scribbles are used for P-Net + CRF, BIFSeg(-w) and BIFSeg. All these methods improve the segmentation. However, some large mis-segmentations can still be observed for P-Net + CRF and BIFSeg(-w). 
In contrast, BIFSeg achieves better results with the same set of scribbles. For a quantitative comparison, we measured the segmentation accuracy after a single round of refinement using the same set of scribbles. The result is shown in Table~\ref{tab:fetal_supervised_seg}. BIFSeg achieves significantly  better accuracy ($p$-value $<$ 0.05) for the placenta, and previously unseen fetal lungs and maternal kidneys compared with P-Net + CRF and BIFSeg(-w). 

\subsubsection{Comparison with Other Interactive Methods} 
The two users (an Obstetrician and a Radiologist) used SlicSeg~\cite{Wang2016}, GrabCut~\cite{Rother2004}, Random Walks~\cite{Grady2005} and BIFSeg for the fetal MRI segmentation tasks respectively. For each image, the user implemented the segmentation interactively until the result was accepted by the user.
The user time and final accuracy of  are presented in Fig.~\ref{fig:fetal_compare_grabcut}. It shows that BIFSeg takes noticeably less user time with similar or higher accuracy compared with the other three interactive segmentation methods.

\subsection{3D Segmentation of Brain Tumors from T1c and FLAIR}\label{sec:3d_segmentation}
\begin{figure}[t]
	\centering 
	\centering
\includegraphics[width=1.0\linewidth]{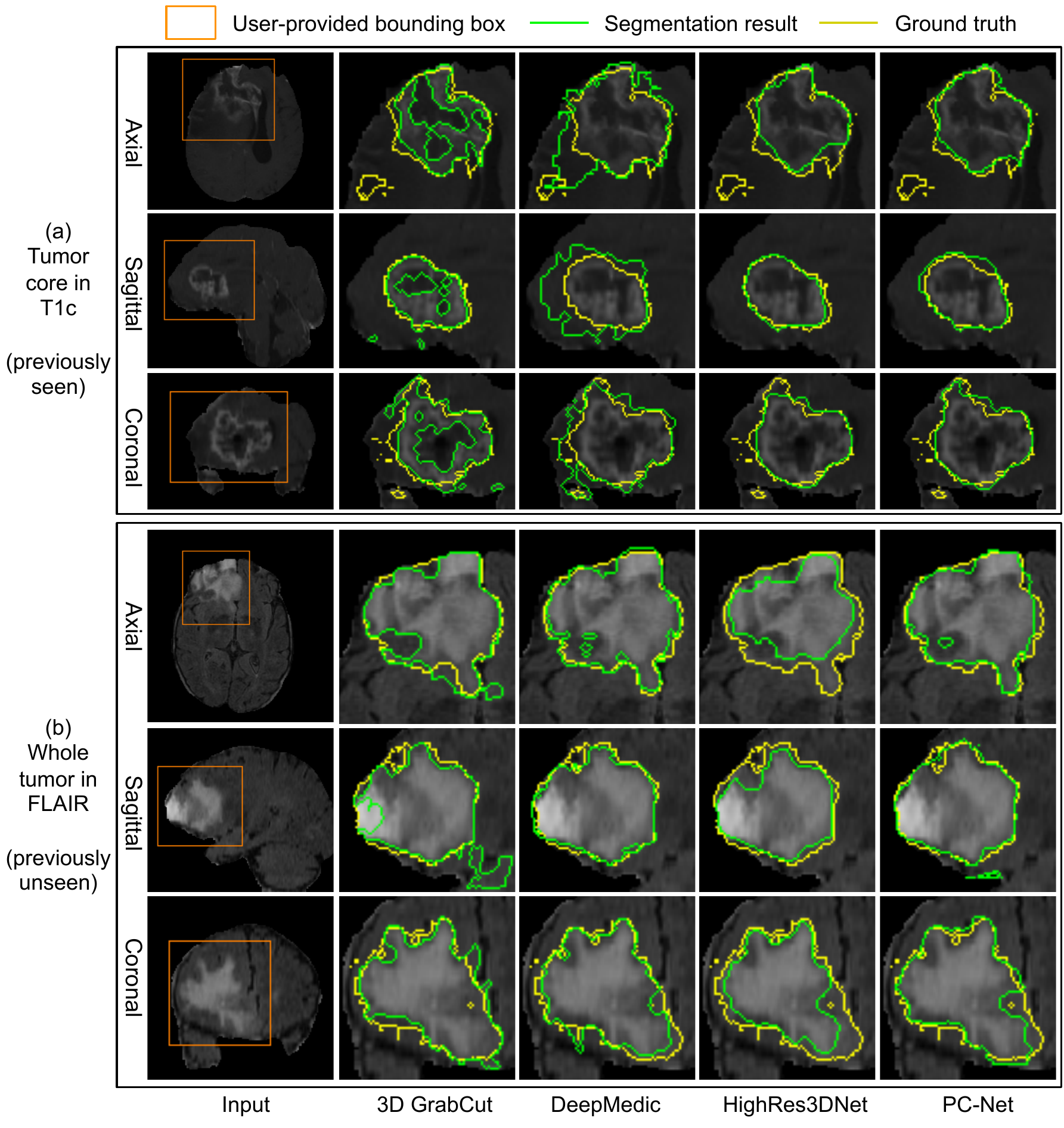}
	\caption[Visual comparison of different segmentation methods with the same bounding]{ 
		Visual comparison of initial segmentation of brain tumors from a 3D bounding box. 
		The whole tumor in FLAIR is previously unseen in the training set. All these methods use the same bounding box for each test image. 
	} 
	\label{fig:different_net_t1c_flair}
\end{figure}
\begin{figure*}[t]
	\centering 
	\centering
\includegraphics[width=0.8\linewidth]{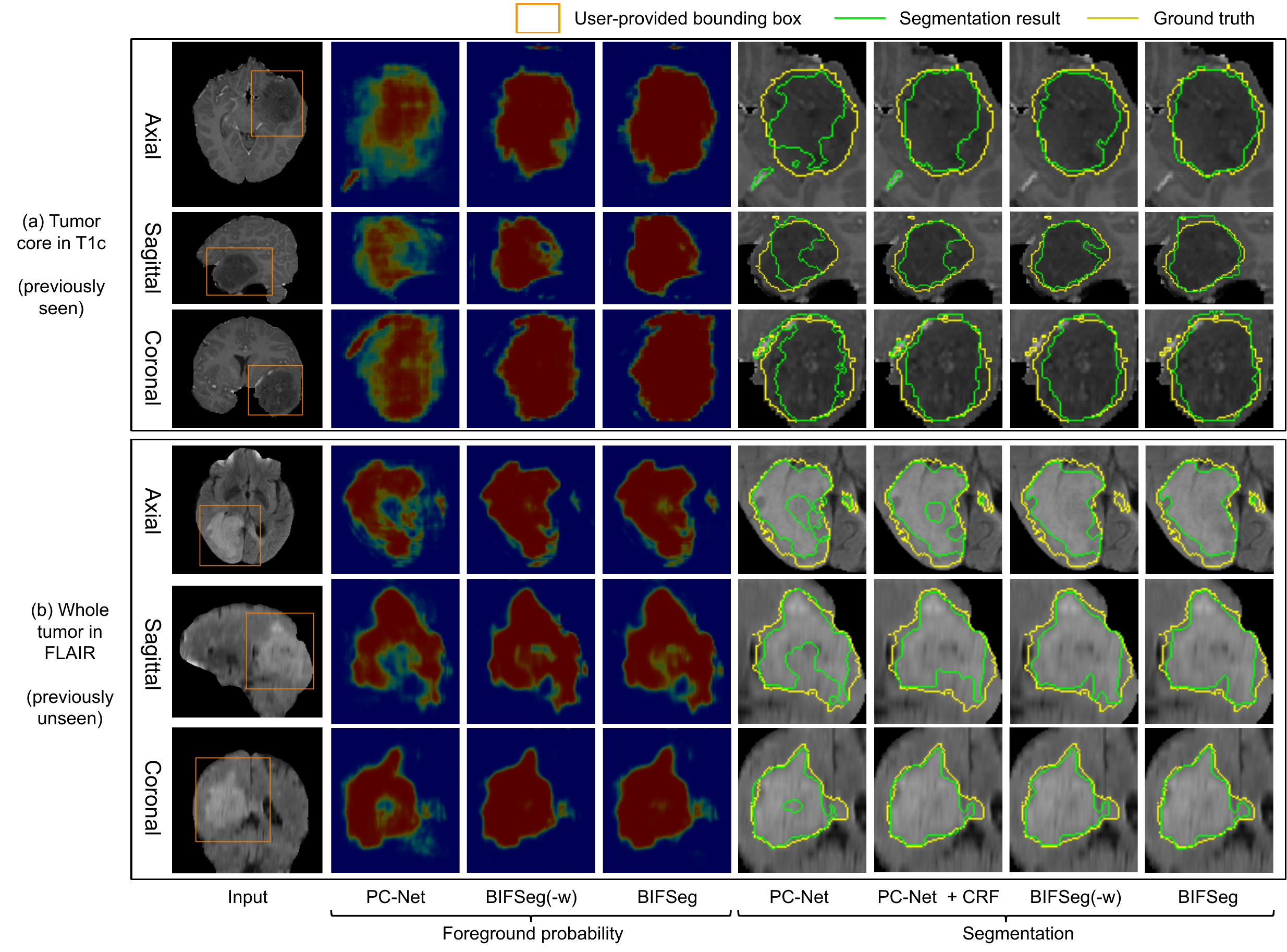}
	\caption[Visual comparison of unsupervised refinement methods without additional scribbles]{ 
		Visual comparison of PC-Net and unsupervised refinement methods without additional scribbles for 3D brain tumor segmentation. The same initial segmentation obtained by PC-Net is used by different refinement methods.
	} 
	\label{fig:tumor_unsupervise}
\end{figure*}

\begin{table}
	\centering
	\footnotesize
	\caption{Dice score of initial segmentation of brain tumors  from a 3D bounding box. 
	All the methods use the same bounding box for each test image. $\wedge$ denotes previously unseen objects. In each row, bold font denotes the best value. * denotes $p$-value $<$ 0.05 compared with the others. 
	}
	\label{tab:different_net}
	\begin{tabular}{l|l|l|l|l}
		\hline
		& DeepMedic & HighRes3DNet & PC-Net & 3D GrabCut \\ \hline
		TC  & 76.68$\pm$11.83 & \bf{83.45$\pm$7.87}  & 82.66$\pm$7.78 & 69.24$\pm$19.20\\ 
		 WT$^{\wedge}$  & \bf{84.04$\pm$8.50} & 75.60$\pm$8.97 & 83.52$\pm$8.76 & 78.39$\pm$18.66 \\ \hline
	\multicolumn{5}{l}{TC: Tumor core in T1c, WT: Whole tumor in FLAIR.} 
	\end{tabular}
\end{table}
\begin{table}
	\centering
	\footnotesize
	\caption{Quantitative comparison of PC-Net and unsupervised refinement methods without additional scribbles for 3D brain tumor segmentation. $T_m$ is the machine time for refinement. $\wedge$ denotes previously unseen objects. In each row, bold font denotes the best value. * denotes $p$-value $<$ 0.05 compared with the others. }
	\label{tab:tumor_unsupervise}
	\begin{tabular}{@{}p{0.55cm}|@{}p{0.7cm}|@{}p{1.3cm}|p{1.55cm}|p{1.35cm}|p{1.35cm}}
		\hline
		&   & ~PC-Net & PC-Net+CRF & BIFSeg(-w) & BIFSeg  \\ \hline
		\multirow{2}{*}{\vtop{\hbox{\strut Dice}\hbox{\strut (\%)}}}
		&  ~TC  & ~82.66$\pm$7.78 & 84.33$\pm$7.32  & 84.67$\pm$7.44 & \bf{86.13$\pm$6.86*}\\ 
		&  ~WT$^{\wedge}$ & ~83.52$\pm$8.76 & 83.92$\pm$7.33 &  83.88$\pm$8.62 & \bf{86.29$\pm$7.31*}  \\ \hline
		\multirow{2}{*}{$T_m$(s)}  & ~TC & ~- & \bf{0.12$\pm$0.04*}  & 3.36$\pm$0.82 & 3.32$\pm$0.82\\
		& ~WT$^{\wedge}$ & ~- & \bf{0.11$\pm$0.05*} & 3.16$\pm$0.89 & 3.09$\pm$0.83\\  
		\hline
		\multicolumn{6}{@{}l}{TC: Tumor core in T1c, WT: Whole tumor in FLAIR.}
	\end{tabular}
\end{table}
\subsubsection{Data}
To validate our method with 3D images, we used the 2015 Brain Tumor Segmentation Challenge (BRATS) training set~\cite{Menze2015}. The ground truth were manually delineated by experts. 
This dataset was collected from 274 cases with multiple MR sequences that give different contrasts. T1c highlights the tumor without peritumoral edema, designated \say{tumor core} as per~\cite{Menze2015}. FLAIR highlights the tumor with peritumoral edema, designated \say{whole tumor} as per~\cite{Menze2015}. We investigate interactive segmentation of tumor cores from T1c images and whole tumors from FLAIR images, which is different from previous works on automatic multi-label and multi-modality segmentation~\cite{Fidon2017a, Kamnitsas2017}. For tumor core segmentation, we randomly selected 249 T1c volumes as our training set and used the remaining 25 T1c volumes as the testing set.  Additionally, to investigate dealing with unseen objects, we employed such trained CNNs to segment whole tumors in the corresponding FLAIR images of these 25 volumes that were not present in our training set. All these images had been skull-stripped and resampled to isotropic 1mm$^3$ resolution. To deal with 3D tumor cores and whole tumors at different scales, we resized the cropped image region inside a bounding box to make its maximal value of width, height and depth be 80. Parameter setting was $\lambda$
= 10.0, $\sigma$ = 0.1, $t_0$ = 0.2, $t_1$ = 0.6, $\epsilon$ = 0.2, $\omega$ = 5.0 based on a grid search with the training data.
\subsubsection{Initial Segmentation based on PC-Net}
 Fig.~\ref{fig:different_net_t1c_flair}(a) shows an initial result of tumor core segmentation from T1c with a user-provided bounding box. Since the central region of the tumor has a low intensity close to that of the background, 3D GrabCut has a poor performance with under-segmentations. DeepMedic leads to some over-segmentations. HighRes3DNet and PC-Net obtain similar results, but PC-Net is less complex and has a lower memory consumption.  Fig.~\ref{fig:different_net_t1c_flair}(b) shows an initial segmentation result of previously unseen whole tumor from FLAIR. 3D GrabCut fails to get high accuracy due to intensity inconsistency in the tumor region, and the CNNs outperform 3D GrabCut, with DeepMedic and PC-Net performing better than HighRes3DNet. A quantitative comparison is presented in Table~\ref{tab:different_net}. It shows that the performance of DeepMedic is low for T1c but high for FLAIR, and that of HighRes3DNet is the opposite. This is because DeepMedic has a small receptive field and tends to rely on local features. It is difficult to use local features to deal with T1c due to its complex appearance but easier to deal with FLAIR since the appearance is less complex. HighRes3DNet has a more complex model and tends to over-fit tumor core. In contrast, PC-Net achieves a more stable performance on tumor core and previously unseen whole tumor. The average machine time for 3D GrabCut, DeepMedic,  and PC-Net is 3.87s, 65.31s and 3.83s, respectively (on the laptop), and that for HighRes3DNet is 1.10s (on the cluster).

\subsubsection{Unsupervised Image-specific Fine-tuning}

Fig.~\ref{fig:tumor_unsupervise} shows unsupervised fine-tuning for brain tumor segmentation based on the initial output of PC-Net without additional user interactions. In Fig.~\ref{fig:tumor_unsupervise}(a), the tumor core is under-segmented in the initial output of PC-Net. CRF improves the segmentation to some degree, but large areas of under-segmentation still exist. The segmentation result of BIFSeg(-w) is similar to that of CRF. In contrast, BIFSeg performs better than CRF and BIFSeg(-w). A similar situation is observed in Fig.~\ref{fig:tumor_unsupervise}(b) for segmentation of previously unseen whole tumor. A quantitative comparison of these methods is shown in Table~\ref{tab:tumor_unsupervise}. BIFSeg improves the average dice score from 82.66\% to 86.13\% for tumor core, and from 83.52\% to 86.29\% for whole tumor. 

\subsubsection{Supervised Image-specific Fine-tuning}
\begin{figure}[t]
	\centering 
	\centering
\includegraphics[width=1.0\linewidth]{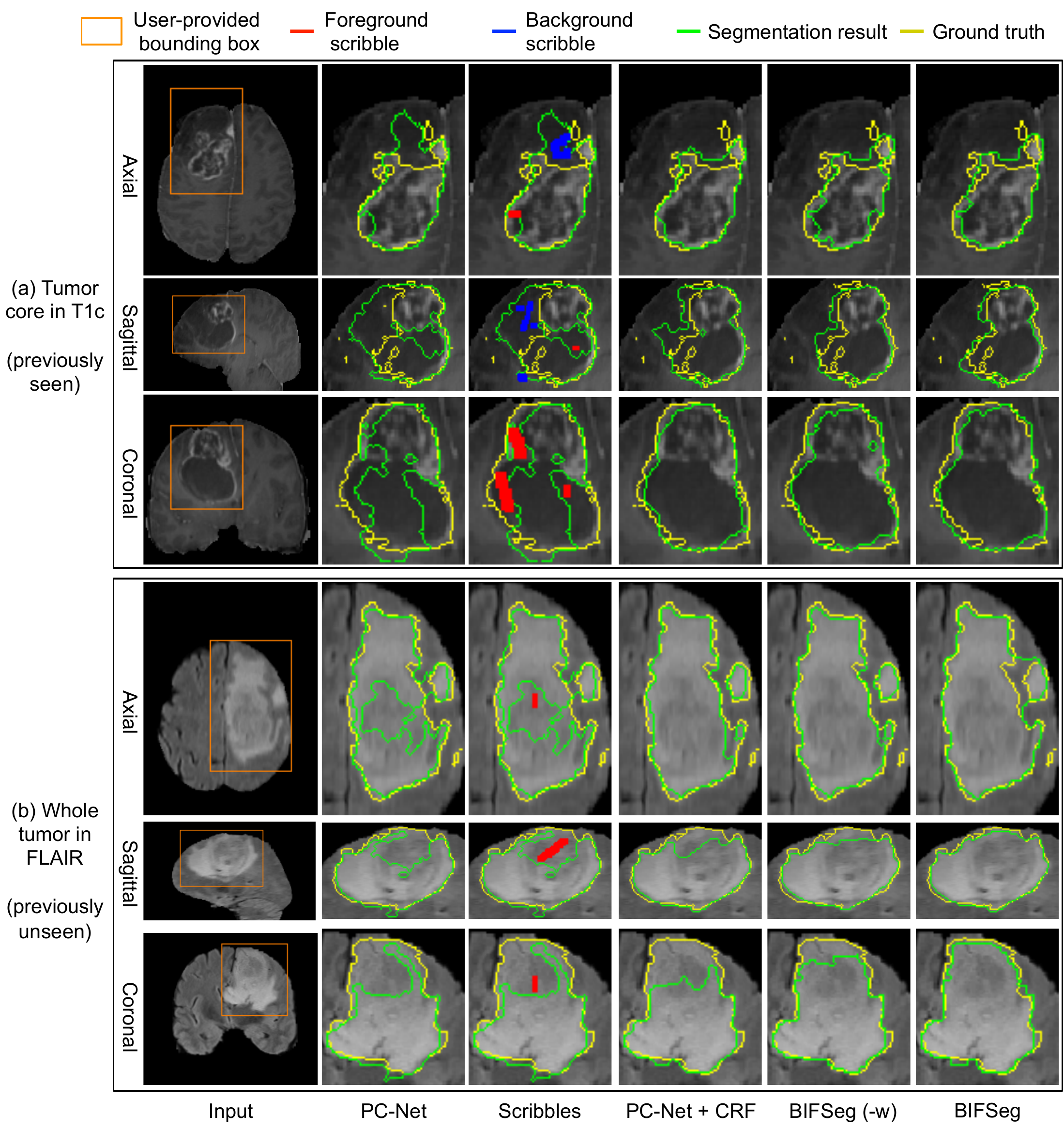}
	\caption[Comparison between different segmentation methods using user-provided bounding box with scribbles]{ 
		Visual comparison of PC-Net and three supervised refinement methods with scribbles for 3D brain tumor segmentation.  The refinement methods use the same initial segmentation and set of scribbles. 
	} 
	\label{fig:tumor_supervise_finetune}
\end{figure}
\begin{table}
	\centering
	\footnotesize
	\caption{Quantitative comparison of PC-Net and three supervised refinement methods with additional scribbles for 3D brain tumor segmentation. $T_m$ is the machine time for refinement. 
		$\wedge$ denotes previously unseen objects. In each row, bold font denotes the best value. * denotes $p$-value $<$ 0.05 compared with the others. }
	\label{tab:tumor_supervise_finetune}
	\begin{tabular}{@{}p{0.55cm}|@{}p{0.7cm}|@{}p{1.3cm}|p{1.55cm}|p{1.35cm}|p{1.35cm}}
		\hline
		&   & ~PC-Net & PC-Net+CRF & BIFSeg(-w) & BIFSeg  \\ \hline
		\multirow{2}{*}{\vtop{\hbox{\strut Dice}\hbox{\strut (\%)}}}
		& ~TC & ~82.66$\pm$7.78 & 85.93$\pm$6.64 & 85.88$\pm$7.53 & \bf{87.49$\pm$6.36*}\\ 
		& ~WT$^{\wedge}$ & ~83.52$\pm$8.76 & 85.18$\pm$6.78 &  86.54$\pm$7.49 & \bf{88.11$\pm$6.09*}  \\ \hline
		\multirow{2}{*}{$T_m$(s)} & ~TC & ~- & \bf{0.14$\pm$0.06*} &  3.33$\pm$0.86 & 4.42$\pm$1.88\\
		&  ~WT$^{\wedge}$ & ~- & \bf{0.12$\pm$0.05*} &  3.17$\pm$0.87 & 4.01$\pm$1.59\\  
		\hline
		\multicolumn{6}{@{}l}{TC: Tumor core in T1c, WT: Whole tumor in FLAIR.}
	\end{tabular}
\end{table}
\begin{figure}[t]
	\centering 
	\centering
\includegraphics[width=1.0\linewidth]{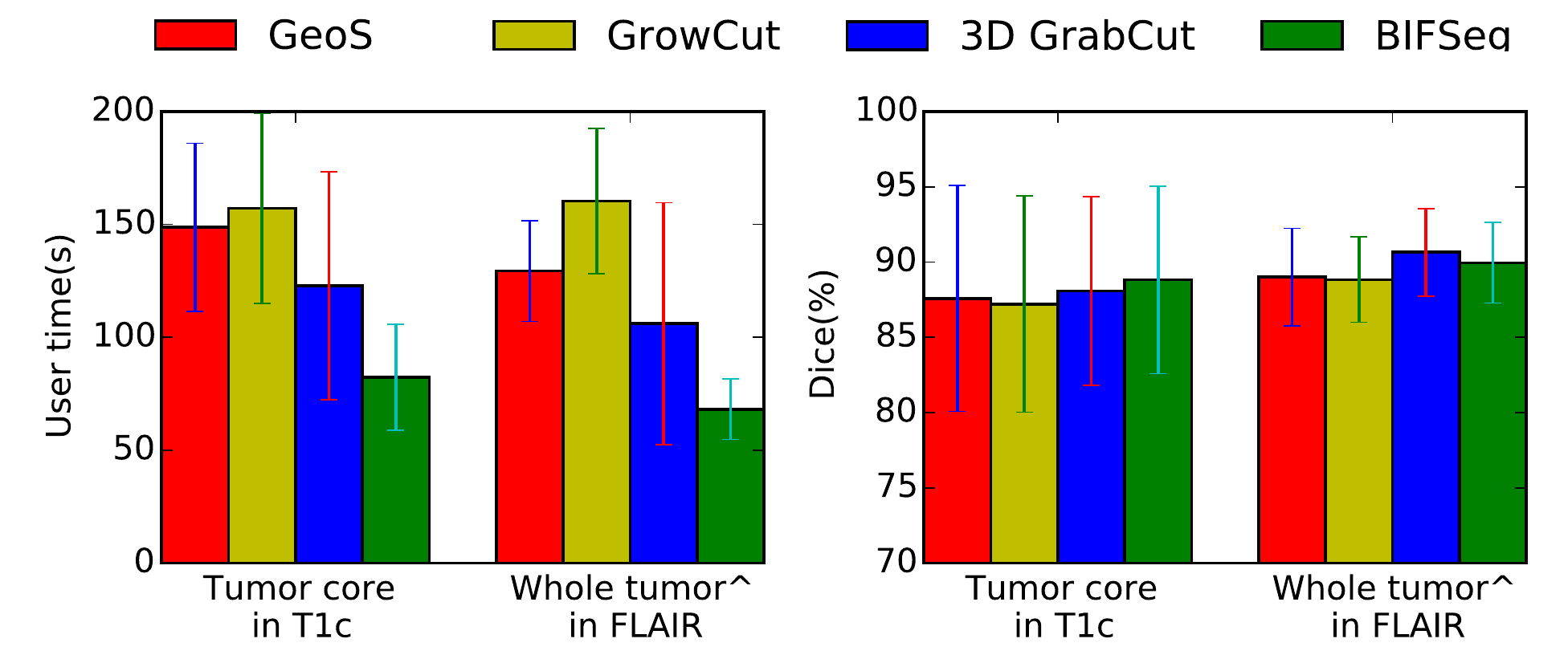}
	\caption[Comparison between different segmentation methods using user-provided bounding box with scribbles]{ 
		User time and Dice score of different interactive methods for 3D brain tumor segmentation. $^\wedge$ denotes previously unseen objects for BIFSeg.
	} 
	\label{fig:tumor_compare_grabcut}
\end{figure}
Fig~\ref{fig:tumor_supervise_finetune} shows refined results of brain tumor segmentation with additional scribbles provided by the user. The same initial segmentation based on PC-Net and the same scribbles are used by CRF, BIFSeg(-w) and BIFSeg. 
It can be observed that CRF and BIFSeg(-w) correct the initial segmentation moderately. In contrast, BIFSeg achieves better refined results for both tumor cores in T1c and whole tumors in FLAIR. For a quantitative comparison of these refinement methods, we measured the segmentation accuracy after a single round of refinement using the same set of scribbles based on the same initial segmentation. The result is shown in Table~\ref{tab:tumor_supervise_finetune}. BIFSeg achieves an average dice score of 87.49\% and 88.11\% for tumor core and previously unseen whole tumor, respectively, and it significantly outperforms CRF and BIFSeg(-w). 
\subsubsection{Comparison with Other Interactive Methods} 
The two users (an Obstetrician and a Radiologist) used GeoS~\cite{Criminisi2008}, GrowCut~\cite{Vezhnevets2005}, 3D GrabCut~\cite{Ram2013} and BIFSeg for the brain tumor segmentation tasks respectively. 
The user time and final accuracy of these methods are presented in Fig.~\ref{fig:tumor_compare_grabcut}. It shows that these interactive methods achieve similar final Dice scores for each task. However, BIFSeg takes significantly less user time to get the results, which is 82.3s and 68.0s in average for tumor core and whole tumor, respectively. 

\section{Discussion and Conclusion}
For 2D images, our P-Net is trained with placenta and fetal brain only, but it performs well on previously unseen fetal lungs and maternal kidneys. For 3D images, the PC-Net is only trained with tumor cores in T1c, but it also achieves good results for whole tumors in FLAIR that are not present for training. This is a major advantage compared with traditional CNNs and even transfer learning~\cite{Tajbakhsh2016} or weakly supervised learning~\cite{Rajchl2016}, since for some objects it does not require annotated instances for training at all. It therefore reduces the efforts needed for gathering and annotating training data and can be applied to some unseen organs directly.
Our proposed framework accepts bounding boxes and optional scribbles as user interactions. Bounding boxes in test images are provided by the user, but they could potentially be obtained by automatic detection~\cite{Keraudren2014} to further increase efficiency. Experimental results show that the image-specific fine-tuning improves the segmentation performance. This acts as a post-processing step after the initial segmentation and outperforms CRF. We found that taking advantage of uncertainty plays an important role for the image-specific fine-tuning process. The uncertainty is defined based on softmax probability and geodesic distance to scribbles if scribbles are given. Recent works~\cite{Gal2016} suggest that test-time dropout also provides classification uncertainty. However, test-time dropout is less suited for interactive segmentation since it leads to longer computational time. 

In conclusion, we propose an efficient deep learning-based framework for interactive 2D/3D medical image segmentation. It uses a bounding box-based CNN for binary segmentation and can segment previously unseen objects. A unified framework is proposed for both unsupervised and supervised refinements of the initial segmentation, where image-specific fine-tuning based on a weighted loss function is proposed. Experiments on segmenting multiple organs from 2D fetal MRI and brain tumors from 3D MRI show that our method performs well on previously unseen objects and the image-specific fine-tuning outperforms CRF. BIFSeg achieves similar or higher accuracy with fewer user interactions in less time than traditional interactive segmentation methods.


%



\section*{Acknowledgment}
This work was supported by the Wellcome Trust (WT101957, WT97914, HICF-T4-275), the EPSRC (NS/A000027/1, EP/H046410/1, EP/J020990/1, EP/K005278, NS/A000050/1), Wellcome/EPSRC [203145Z/16/Z], the Royal Society [RG160569], the National Institute for Health Research University College London Hospitals Biomedical Research Centre (NIHR BRC UCLH/UCL), a UCL ORS and GRS, hardware donated by NVIDIA, and by Emerald, a GPU-accelerated High Performance Computer, made available by the Science \& Engineering South Consortium operated in partnership with the STFC Rutherford-Appleton Laboratory.

\ifCLASSOPTIONcaptionsoff
  \newpage
\fi



\bibliographystyle{IEEEtran}
\bibliography{./reference/miccai2017.bib}

\begin{thebibliography}{10}
\providecommand{\url}[1]{#1}
\csname url@samestyle\endcsname
\providecommand{\newblock}{\relax}
\providecommand{\bibinfo}[2]{#2}
\providecommand{\BIBentrySTDinterwordspacing}{\spaceskip=0pt\relax}
\providecommand{\BIBentryALTinterwordstretchfactor}{4}
\providecommand{\BIBentryALTinterwordspacing}{\spaceskip=\fontdimen2\font plus
\BIBentryALTinterwordstretchfactor\fontdimen3\font minus
  \fontdimen4\font\relax}
\providecommand{\BIBforeignlanguage}[2]{{%
\expandafter\ifx\csname l@#1\endcsname\relax
\typeout{** WARNING: IEEEtran.bst: No hyphenation pattern has been}%
\typeout{** loaded for the language `#1'. Using the pattern for}%
\typeout{** the default language instead.}%
\else
\language=\csname l@#1\endcsname
\fi
#2}}
\providecommand{\BIBdecl}{\relax}
\BIBdecl

\bibitem{Litjens2017}
G.~Litjens, T.~Kooi, B.~E. Bejnordi, A.~A.~A. Setio, F.~Ciompi, M.~Ghafoorian,
  J.~A. van~der Laak, B.~V. Ginneken, and C.~I. S{\'{a}}nchez, ``{A survey on
  deep learning in medical image analysis},'' \emph{Medical Image Analysis},
  vol.~42, pp. 60--88, 2017.

\bibitem{Zhao2013}
F.~Zhao and X.~Xie, ``{An overview of interactive medical image
  segmentation},'' \emph{Annals of the BMVA}, vol. 2013, no.~7, pp. 1--22,
  2013.

\bibitem{Grady2005}
L.~Grady, T.~Schiwietz, S.~Aharon, and R.~Westermann, ``{Random walks for
  interactive organ segmentation in two and three dimensions: implementation
  and validation},'' in \emph{MICCAI}, 2005, pp. 773--780.

\bibitem{Criminisi2008}
A.~Criminisi, T.~Sharp, and A.~Blake, ``{GeoS: Geodesic image segmentation},''
  in \emph{ECCV}, 2008, pp. 99--112.

\bibitem{Rajchl2016}
M.~Rajchl, M.~Lee, O.~Oktay, K.~Kamnitsas, J.~Passerat-Palmbach, W.~Bai,
  M.~Rutherford, J.~Hajnal, B.~Kainz, and D.~Rueckert, ``{DeepCut: Object
  segmentation from bounding box annotations using convolutional neural
  networks},'' \emph{TMI}, vol.~36, no.~2, pp. 674--683, 2017.

\bibitem{Xu2016}
N.~Xu, B.~Price, S.~Cohen, J.~Yang, and T.~Huang, ``{Deep interactive object
  selection},'' in \emph{CVPR}, 2016, pp. 373--381.

\bibitem{Wang2017pami}
G.~Wang, M.~A. Zuluaga, W.~Li, R.~Pratt, P.~A. Patel, M.~Aertsen, T.~Doel,
  M.~Klusmann, A.~L. David, J.~Deprest, S.~Ourselin, and T.~Vercauteren,
  ``{DeepIGeoS: A deep interactive geodesic framework for medical image
  segmentation},'' \emph{arXiv preprint arXiv:1707.00652}, 2017.

\bibitem{Ribeiro2006}
H.~L. Ribeiro and A.~Gonzaga, ``{Hand image segmentation in video sequence by
  GMM: a comprarative analysis},'' in \emph{SIBGRAPI}, 2006, pp. 357--364.

\bibitem{Kamnitsas2017}
K.~Kamnitsas, C.~Ledig, V.~F.~J. Newcombe, J.~P. Simpson, A.~D. Kane, D.~K.
  Menon, D.~Rueckert, and B.~Glocker, ``{Efficient multi-scale 3D CNN with
  fully connected CRF for accurate brain lesion segmentation},'' \emph{Medical
  Image Analysis}, vol.~36, pp. 61--78, 2017.

\bibitem{Li2017}
W.~Li, G.~Wang, L.~Fidon, S.~Ourselin, M.~J. Cardoso, and T.~Vercauteren, ``{On
  the compactness, efficiency, and representation of 3D convolutional networks:
  brain parcellation as a pretext task},'' in \emph{IPMI}, 2017, pp. 348--360.

\bibitem{Long2014}
J.~Long, E.~Shelhamer, and T.~Darrell, ``{Fully convolutional networks for
  semantic segmentation},'' in \emph{CVPR}, 2015, pp. 3431--3440.

\bibitem{Chen2015iclr}
L.-C. Chen, G.~Papandreou, I.~Kokkinos, K.~Murphy, and A.~L. Yuille,
  ``{Semantic image segmentation with deep convolutional nets and fully
  connected CRFs},'' in \emph{ICLR}, 2015.

\bibitem{Hefny2015a}
O.~Ronneberger, P.~Fischer, and T.~Brox, ``{U-Net: Convolutional networks for
  biomedical image segmentation},'' in \emph{MICCAI}, 2015, pp. 234--241.

\bibitem{chen_hao2016_dcan}
H.~Chen, X.~Qi, L.~Yu, and P.-A. Heng, ``{DCAN: Deep contour-aware networks for
  accurate gland segmentation},'' in \emph{CVPR}, 2016, pp. 2487--2496.

\bibitem{RichardMckinley2016}
R.~Mckinley, R.~Wepfer, T.~Gundersen, F.~Wagner, A.~Chan, R.~Wiest, and
  M.~Reyes, ``{Nabla-net: A deep dag-like convolutional architecture for
  biomedical image segmentation},'' in \emph{BrainLes}, 2016, pp. 119--128.

\bibitem{Roth2015}
H.~R. Roth, L.~Lu, A.~Farag, H.-c. Shin, J.~Liu, E.~B. Turkbey, and R.~M.
  Summers, ``{DeepOrgan: Multi-level deep convolutional networks for automated
  pancreas segmentation},'' in \emph{MICCAI}, 2015, pp. 556--564.

\bibitem{Milletari2016}
F.~Milletari, N.~Navab, and S.-A. Ahmadi, ``{V-Net: Fully convolutional neural
  networks for volumetric medical image segmentation},'' in \emph{IC3DV}, 2016,
  pp. 565--571.

\bibitem{Dou2017}
Q.~Dou, L.~Yu, H.~Chen, Y.~Jin, X.~Yang, J.~Qin, and P.-A. Heng, ``{3D deeply
  supervised network for automated segmentation of volumetric medical
  images},'' \emph{Medical Image Analysis}, vol.~41, pp. 40--54, 2017.

\bibitem{Boykov2001}
Y.~Y. Boykov and M.~P. Jolly, ``{Interactive graph cuts for optimal boundary
  {\&} region segmentation of objects in N-D images},'' in \emph{ICCV}, 2001,
  pp. 105--112.

\bibitem{Rother2004}
C.~Rother, V.~Kolmogorov, and A.~Blake, ``{GrabCut: Interactive foreground
  extraction using iterated graph cuts},'' \emph{ACM Trans. on Graphics},
  vol.~23, no.~3, pp. 309--314, 2004.

\bibitem{Wang2016}
G.~Wang, M.~A. Zuluaga, R.~Pratt, M.~Aertsen, T.~Doel, M.~Klusmann, A.~L.
  David, J.~Deprest, T.~Vercauteren, and S.~Ourselin, ``{Slic-Seg: A minimally
  interactive segmentation of the placenta from sparse and motion-corrupted
  fetal MRI in multiple views},'' \emph{Medical Image Analysis}, vol.~34, pp.
  137--147, 2016.

\bibitem{Top2011a}
A.~Top, G.~Hamarneh, and R.~Abugharbieh, ``{Active learning for interactive 3D
  image segmentation},'' in \emph{MICCAI}, 2011, pp. 603--610.

\bibitem{Lin2016}
D.~Lin, J.~Dai, J.~Jia, K.~He, and J.~Sun, ``{ScribbleSup: Scribble-supervised
  convolutional networks for semantic segmentation},'' in \emph{CVPR}, 2016,
  pp. 3159--3167.

\bibitem{Abdulkadir2016}
A.~Abdulkadir, S.~S. Lienkamp, T.~Brox, and O.~Ronneberger, ``{3D U-Net :
  Learning dense volumetric segmentation from sparse annotation},'' in
  \emph{MICCAI}, 2016, pp. 424--432.

\bibitem{Tajbakhsh2016}
N.~Tajbakhsh, J.~Y. Shin, S.~R. Gurudu, R.~T. Hurst, C.~B. Kendall, M.~B.
  Gotway, and J.~Liang, ``{Convolutional neural networks for medical image
  analysis: full training or fine tuning?}'' \emph{TMI}, vol.~35, no.~5, pp.
  1299--1312, 2016.

\bibitem{Wang17brats}
G.~Wang, W.~Li, S.~Ourselin, and T.~Vercauteren, ``{Automatic brain tumor
  segmentation using cascaded anisotropic convolutional neural networks},''
  \emph{arXiv preprint arXiv:1709.00382}, 2017.

\bibitem{Jia2014}
Y.~Jia, E.~Shelhamer, J.~Donahue, S.~Karayev, J.~Long, R.~Girshick,
  S.~Guadarrama, and T.~Darrell, ``{Caffe: Convolutional architecture for fast
  feature embedding},'' in \emph{ACMICM}, 2014, pp. 675--678.

\bibitem{Vezhnevets2005}
V.~Vezhnevets and V.~Konouchine, ``{GrowCut: Interactive multi-label ND image
  segmentation by cellular automata},'' in \emph{Graphicon}, 2005, pp.
  150--156.

\bibitem{Ram2013}
E.~Ram and P.~Temoche, ``{A volume segmentation approach based on GrabCut},''
  \emph{CLEI Electronic Journal}, vol.~16, no.~2, pp. 4--4, 2013.

\bibitem{Menze2015}
B.~H. Menze, A.~Jakab, S.~Bauer, J.~Kalpathy-Cramer, K.~Farahani, J.~Kirby,
  Y.~Burren, N.~Porz, J.~Slotboom, R.~Wiest, L.~Lanczi, E.~Gerstner, M.~A.
  Weber, T.~Arbel, B.~B. Avants, N.~Ayache, P.~Buendia, D.~L. Collins,
  N.~Cordier, J.~J. Corso, A.~Criminisi, T.~Das, H.~Delingette,
  {\c{C}}.~Demiralp, C.~R. Durst, M.~Dojat, S.~Doyle, J.~Festa, F.~Forbes,
  E.~Geremia, B.~Glocker, P.~Golland, X.~Guo, A.~Hamamci, K.~M. Iftekharuddin,
  R.~Jena, N.~M. John, E.~Konukoglu, D.~Lashkari, J.~A. Mariz, R.~Meier,
  S.~Pereira, D.~Precup, S.~J. Price, T.~R. Raviv, S.~M. Reza, M.~Ryan,
  D.~Sarikaya, L.~Schwartz, H.~C. Shin, J.~Shotton, C.~A. Silva, N.~Sousa,
  N.~K. Subbanna, G.~Szekely, T.~J. Taylor, O.~M. Thomas, N.~J. Tustison,
  G.~Unal, F.~Vasseur, M.~Wintermark, D.~H. Ye, L.~Zhao, B.~Zhao, D.~Zikic,
  M.~Prastawa, M.~Reyes, and K.~{Van Leemput}, ``{The multimodal brain tumor
  image segmentation benchmark (BRATS)},'' \emph{TMI}, vol.~34, no.~10, pp.
  1993--2024, 2015.

\bibitem{Fidon2017a}
L.~Fidon, W.~Li, L.~C. Garcia-Peraza-Herrera, J.~Ekanayake, N.~Kitchen,
  S.~Ourselin, and T.~Vercauteren, ``{Scalable multimodal convolutional
  networks for brain tumour segmentation},'' in \emph{MICCAI}, 2017, pp.
  285--293.

\bibitem{Keraudren2014}
K.~Keraudren, M.~Kuklisova-Murgasova, V.~Kyriakopoulou, C.~Malamateniou, M.~A.
  Rutherford, B.~Kainz, J.~V. Hajnal, and D.~Rueckert, ``{Automated fetal brain
  segmentation from 2D MRI slices for motion correction},'' \emph{NeuroImage},
  vol. 101, pp. 633--643, 2014.

\bibitem{Gal2016}
Y.~Gal and Z.~Ghahramani, ``{Dropout as a bayesian approximation: representing
  model uncertainty in deep learning},'' in \emph{ICML}, 2016, pp. 1050--1059.

\end{thebibliography}
\end{document}